\documentclass[letterpaper, 10 pt, conference]{ieeeconf}  

\makeatletter
\def\endthebibliography{%
	\def\@noitemerr{\@latex@warning{Empty `thebibliography' environment}}%
	\endlist
}
\makeatother

\IEEEoverridecommandlockouts                              

\usepackage{graphicx}
\usepackage{amsmath} 
\usepackage{subfig}
\usepackage{url}
\usepackage{amssymb}  

\newcommand{\M}[1]{\mathbf{#1}}
\newcommand{\V}[1]{\mathbf{#1}}

\usepackage[
    bookmarks=true,          
    bookmarksopen=true,      
    bookmarksopenlevel=1,    
    bookmarksnumbered=true,  
    unicode=false,           
    pdftoolbar=true,         
    pdfmenubar=true,         
    pdffitwindow=false,      
    pdfstartview={FitH},     
    pdftitle={},             
    pdfauthor={},            
    pdfsubject={},           
    colorlinks=true,         
    linkcolor=red,           
    citecolor=green,         
    filecolor=magenta,       
    urlcolor=cyan,           
    breaklinks=true,    
    plainpages=false,   
    pdfpagelabels       
]{hyperref}

\newcommand{\columnname}[1]
{\makebox[\tempwidth][c]{\textbf{#1}}}

\title{Non-iterative One-step Solution for Point Set Registration Problem\\ on Pose Estimation without Correspondence
}

\author{Yijun Yuan$^{1}$, Dorit Borrmann$^{2}$, Andreas N\"uchter$^{2}$ and S\"oren Schwertfeger$^{1}$
	\thanks{$^{1}$Yijun Yuan and S\"oren Schwertfeger with the School of Information Science and Technology, 
		ShanghaiTech University, China.
		{\tt\small [yuanwj|soerensch]@shanghaitech.edu.cn}}%
	\thanks{$^{2}$Dorit Borrmann, Andreas N\"uchter are with the Department of Informatics VII -- Robotics and Telematics, 
		Julius-Maximilians-University W\"urzburg, Germany.}
}

\begin{document}

\maketitle
\thispagestyle{empty}
\pagestyle{empty}

\begin{abstract}
	In this work, we propose to directly find the one-step solution for
	the point set registration problem without correspondences. Inspired
	by the Kernel Correlation method, we consider the fully connected
	objective function between two point sets, thus avoiding the
	computation of correspondences. By utilizing least square minimization,
	the transformed objective function is directly solved with existing
	well-known closed-form solutions, e.g., singular value decomposition,
	that is usually used for given correspondences. However, using equal
	weights of costs for each connection will degenerate the solution due
	to the large influence of distant pairs. Thus, we additionally set a
	scale on each term to avoid  high costs on non-important pairs. As
	in feature-based registration methods, the similarity between descriptors of points
	determines the scaling weight. Given the weights, we get a one step
	solution. As the runtime is in $\mathcal O (n^2)$, we also propose a
	variant with keypoints that strongly reduces the cost. The experiments
	show that the proposed method gives a one-step solution without an
	initial guess. Our method exhibits competitive outlier robustness and accuracy, compared to various other
	methods, and it is more stable in case of large rotations. Additionally, our one-step solution achieves a performance on-par with the state-of-the-art feature based method TEASER.
\end{abstract}

\section{Introduction}
The point set registration problem has been explored for several
decades. Various techniques have been invented, focusing on both
efficiency and accuracy.
%
As discussed in~\cite{li20073d}, it is extremely hard to find the
optimal transformation $\M T$ and correspondence matrix $\M P$
simultaneously. The problem has been addressed in~\cite{li20073d} by
alternating the optimization of $\M T$ and $\M P$.

In recent decades, a multitude of algorithms have been proposed on 3D
registration. They are divided into rigid and non-rigid
algorithms~\cite{bellekens2014survey} and work either iteratively
to solve for the transformation matrix with repeatedly matched
points~\cite{marden2012improving,besl1992method,fantoni2012accurate,segal2009generalized,rusu2009fast}
or treat the problem as an optimization program that omits the
necessity of computing correspondences
\cite{tsin2004correlation,myronenko2010point,zheng2009fast}.
With the high capability of regression methods for Deep Neural
Networks, there are some attempts to directly solve the transformation
with deep neural networks. Researchers start to seek for approaches that
directly predict the transformation~\cite{miao2016cnn}.
However, those trained models highly rely on the learning data that
make it both very costly and not reliable to cases that are not
covered by the space of training data.

\begin{figure}
	\centering
	\includegraphics[width=.75\linewidth]{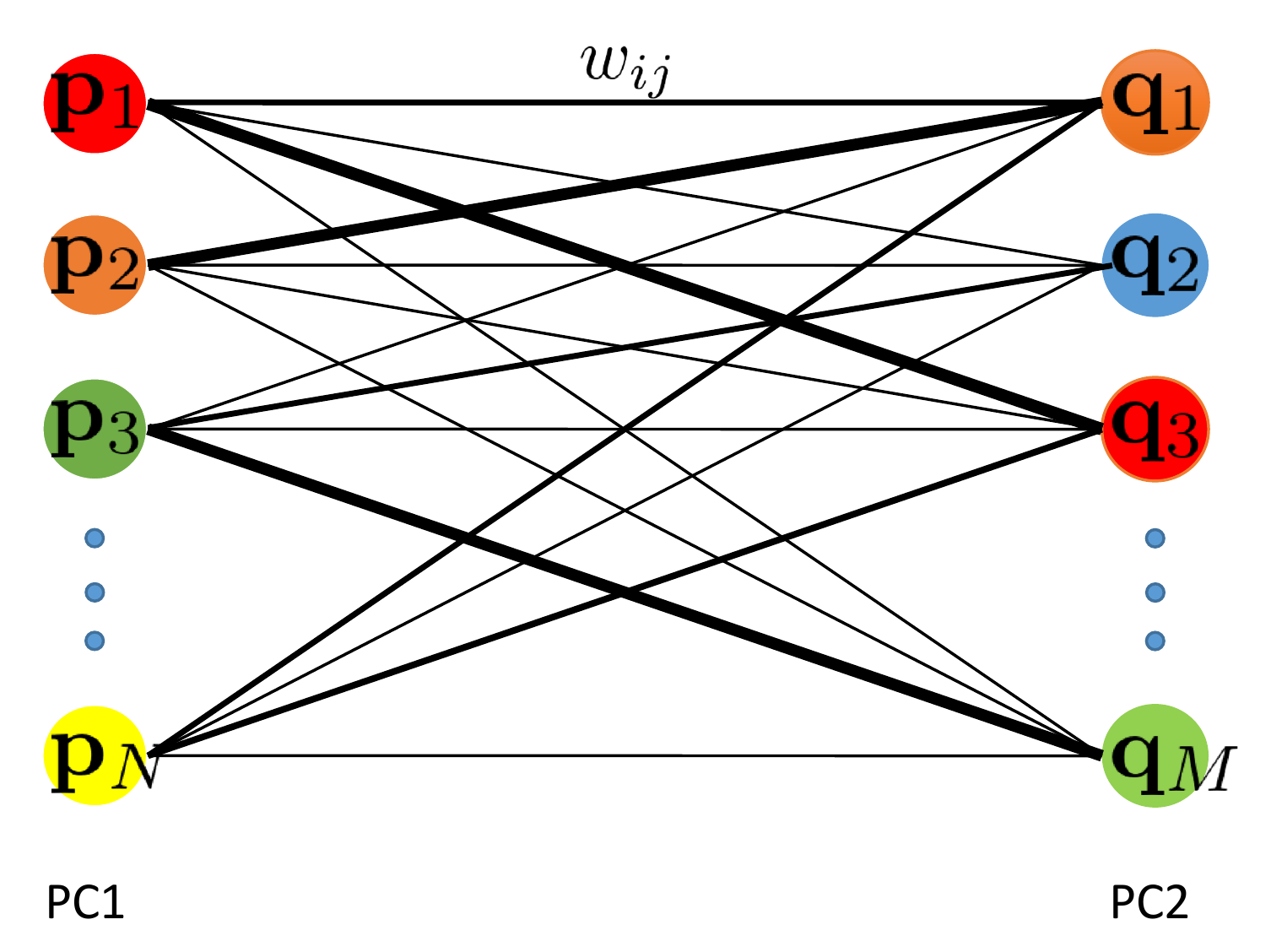}		
	\caption{Full connection between two point sets. Each edge is a
		weighted Euclidean squared distance term in our object
		function, given a proper $w_{i,j}$ to scale the cost term of
		the pair $(i,j)$. The thickness of the lines reflect the similarity (weight) of pairs.}
	\label{fig:fc}
     \vspace*{-4mm}
\end{figure}

This paper presents a direct solution to the point cloud registration
problem without the need of a trained model.
There are two problems to address: Correspondence computation and
optimization of the objective function.
Kernel correlation (KC)~\cite{tsin2004correlation} is one of the most
common registration methods that solves the problem without known
correspondences by minimizing the full connection cost between two
point sets. It is generally in the form $-\mathcal K(\mathbf X,\mathbf Y)$
with $\mathcal K$ being the kernel.
Inspired by the distribution distance, we consider the full connection
loss as a good way to omit the correspondence computation.

Aiming at a one-step solution, we first review closed-form
results. Given correspondences, there are four known
possibilities~\cite{Walker_1991,Horn_1987,Horn_1988,Arun_1987}. The
singular value decomposition (SVD) is widely used for computing the
optimal rotation $\M R$ and afterwards the optimal translation $\V
t$~\cite{Arun_1987}. Our full connection function builds on this
least square solution.

But if each cost term has an equivalent effect on the objective function, the
method will fail. In the KC method, the costs between very distant
points only has a tiny impact due to its kernel function. However, in
the least square case, large distances will dominate the
system and thus do not perform well. 
Therefore, we properly weigh each cost term to suppress the influence
of distant point pairs. The weights consider the similarity between
two points. Fig.~\ref{fig:fc} illustrates the full connection, 
weights are set according to a similarity measure.

In the following, we first formulate the problem and show some related
solutions. Then our method is detailed in
Section~\ref{sec::method}. After that, in Section \ref{sec::exp}, we
present the experiments for sensitivity to noise, robustness to outliers,
and overall accuracy.

\section{Related Work}
\label{sec::relatedWork}

The Iterative Closest Point (ICP) algorithm is the most famous
registration method. It has been widely applied to various
representations of 3D shapes~\cite{besl1992method} and is able to
align a set of range images into a 3D
model~\cite{fantoni2012accurate}. The
generalized-ICP~\cite{segal2009generalized} even puts point-to-point
ICP and point-to-plane ICP into one probabilistic framework.
ICP consists of two steps, correspondence search and solving
for the optimal transformation.

To speed up the search process, point clouds often are stored in
$k$-d trees. To make it faster, Marden and Guivant
\cite{marden2012improving} propose to use a grid data structure to
provide a constant time approximate nearest neighbor search. To achieve
better quality for matching, especially of very large point clouds,
feature based methods are used. Fast Point Feature Histograms (FPFH) are
used to analyze the local geometry around a 3D point and provide a
basis for the fast computation of a descriptor~\cite{rusu2009fast}.

Given known correspondences, the transformation can be computed. Walker et
al. use dual number quaternions and formulate it as an optimization
problem~\cite{Walker_1991}. With a matrix of sum of products of
corresponding point coordinates, Horn computes the optimal rotation
from the eigenvector associated to the largest positive
eigenvalue~\cite{Horn_1987}. This eigenvector is a unit quaternion
representing the rotation.

However, the least square form using a matrix representation of rotation
is more common. The problem is formulated as follows: Assume we have
two point clouds $\mathbf P$ and $\mathbf Q$ with $\V p_i\in
\mathbf P|_{i\in \{1,\cdots N\}}$ and $\V q_j\in \mathbf Q |_{j\in
	\{1,\cdots M\}}$. Since we have 3D point clouds, $\V p_i,\V
q_j\in \mathbb R^3$.
Then the optimization task is
\begin{align}
\label{eq:loss1}
\min_{\M R, \V t}\sum_{(i,j)\in \mathcal C}||  \M R\V p_i + \V t - \V q_i   ||^2 
\end{align}
where $\M R$, $\V t$ are the rotation matrix and translation vector to
transform $\mathbf P$ into the coordinate system of $\mathbf
Q$. $\mathcal C$ is the set of correspondences.

With a more widely used orthogonal matrix representation for the
rotation, Horn et al. \cite{Horn_1988} formulate a least square problem
and propose a solution using a 3-by-3 matrix.
Also relying on a matrix representation, Arun et al.~\cite{Arun_1987}
resort to the SVD to solve for the rotation as a multiplication of the
two resulting orthonormal matrices.

However, in ICP and related methods, the correspondences have to be
recomputed each iteration. To avoid this, the KC method
\cite{tsin2004correlation} uses an objective function that fully
connects the point clouds:
\begin{align}
\label{eq:loss2}
\min_{\M R, \V t}\sum_{i=1}^{N}\sum_{j=1}^{M} e^{-\frac{|| \M R\V p_i
		+ \V t - \V q_j ||^2}{2\sigma^2} }
\end{align}
if Gaussian distances are chosen. In each term of the summation, a
robust function, the Gaussian distance, has been utilized.  Similar to
Maximum Mean Discrepancy (MMD), KC evaluates the distance between two
distributions. Thus it shows better sensitivity to noise and is more
robust than ICP-like methods. Some recent publications do not rely
on correspondences. Myronenko and Song~\cite{myronenko2010point}
represent point clouds with Gaussian mixture models and solve the
transformation by aligning the model centroids. Zheng et
al.~\cite{zheng2009fast} build a continuous distance field for a fixed
model and align the other point set model to minimize the energy
iteratively. Yang et al.~\cite{yang2019polynomial} reformulate the registration as a truncated least squares estimation (TEASER) which is thus robust with extremely wrong correspondences. 

\section{Methodology}
\label{sec::method}

Actually, both Equations~\eqref{eq:loss1} and \eqref{eq:loss2} have
their benefits. While needing to compute the correspondences,
Eq.~\eqref{eq:loss1} has a closed-form solution. The KC loss
Eq.~\eqref{eq:loss2} omits the necessity of finding the
correspondences.

We intend to use both full connection and the least square
form. However, just replacing the kernel with the quadratic distance
will not work due to the distant pairs that will dominate the loss. As
discussed in \cite{tsin2004correlation}, the gradient of the quadratic
function is very sensitive to outliers, so a more robust function, the
Gaussian kernel, has been utilized.
To avoid the fast increase of the gradient, we use additional weights
to rebalance each quadratic term in the full connection.
%
The formula is a summation of square distances for each fully
connected point pair. The weight $w_{i,j}$ in the range $(0,1]$ has
been assigned for each term.
\begin{align}
\label{eq:loss3}
\min_{\M R, \V t}\sum_{i=1}^{N}\sum_{j=1}^{M} w_{i,j}|| \M R\V p_i + \V t - \V q_j   ||^2 
\end{align}
Please note that one problem of Gaussian kernel distances in the KC
method is, that $\sigma$ has to be properly set according to the
scale of the data. We use the square distance, as it is invariant
to scale \cite{fleuret2003scale}. 
However, to have the desired suppression effect, weights cannot
be arbitrarily chosen. We will discuss the weights in
Section~\ref{sec::weight}.

\subsection{Solving the Transformation}

For the weighted function~\eqref{eq:loss3}, there is a full connection
with quadratic distance between every point $\V p\in \mathbf P$ and
$\V q\in \mathbf Q$.
Then the problem is to reformulate Eq.~\eqref{eq:loss3} with full
connection as correspondences. The new point sets
$(\mathcal{X},\mathcal{Y})$ are of size $NM$ and each pair is a
connection.

The optimal solution is obtained with any algorithm that computes the
transformation. To make the paper self-contained, we choose the SVD~\cite{Arun_1987}, also detailed in \cite{sorkine2009least}. Let 
$\mathcal{X}=\{\V p'_1,\cdots \V
p'_{NM} \}$, $\mathcal{Y}=\{\V q'_1,\cdots \V q'_{NM} \}$,
the problem is formulated as
\newcommand{\argmin}{\operatornamewithlimits{argmin}}
\begin{align}
(\M R,\V t) = \argmin_{\M R\in SO(d), \V t\in \mathbb R^d} \sum_{i=1}^{NM} w_i||(\M R \V p'_i+\V t)-\V q'_i ||^2
\end{align}
with known weights $w_i>0$. 

We cancel $\V t$ by computing the weighted mean
\begin{align}
\bar{\V p}' = \frac{\sum_{i=1}^{NM}w_i\V p'_i}{\sum_{i=1}^{NM}w_i}, \
\bar{\V q}' = \frac{\sum_{i=1}^{NM}w_i\V q'_i}{\sum_{i=1}^{NM}w_i}
\end{align}
and centering the point clouds
\begin{align}
\V x_i :=  \V p'_i-\bar{\V p}',  \V y_i :=  \V q'_i-\bar{\V q}' .
\end{align}
We can then compute $\V R$
\begin{align}
\M R = \argmin_{\M R\in SO(d)}\sum_{i=1}^{NM}w_i ||\M R \V x_i- \V y_i  ||^2 .
\end{align}
Let $\M X$ denote the matrix where $\V x_i$ is the $i-th$
column. Similarly, we have $\M Y$. Thus, $\M X,\M Y \in \mathbb
R^{3\times NM}$. $\M W$ is a diagonal matrix with $\M W_{i,i}=w_i$.
The SVD solves it where
\begin{align}
\M U\M \Sigma\M V^T &= \M X\M W\M Y^T
\intertext{and the optimal rotation is computed by}
\M R &= \M V \M U^T.
\end{align} When the solution consists of a reflection, i.e., $|\M V| |\M U|<0$, the last column of $\M V$ will be multiplied with $-1$ before computing the rotation.

Finally, the translation is given as
\begin{align}
\V t =  \bar{\V q}^{'} - \M R \bar{\V p}^{'}.
\end{align}

\subsection{Weights as Similarity of Feature}
\label{sec::weight}

To determine the weights, we use $f_{\mathcal{X}}(\V x)$ to denote a
function that extracts a feature descriptor of the point $\V x$ from
the point cloud $\mathcal{X}$.
Then the similarity is obtained as
\begin{align}
\label{eq:similarity}
w_i = e^{-\frac{1}{\beta}||f_{\mathcal{X}}(\V p'_i) - f_{\mathcal Y}(\V q'_i)||^2}.
\end{align}
The lower the similarity, the lower the weight of the pairs. Thus, the
effect of the term on the objective function will be less. In this
way, a pair of points with low similarity contributes only a little, as
they have a large feature descriptor distance.
The constant $\beta$ in Eq.~\eqref{eq:similarity} scales the feature
distance. It depends on the selected feature descriptor.
We utilize the FPFH \cite{rusu2009fast} for $f$ in our implementation.

\begin{figure}
	\centering
	\includegraphics[width=.95\linewidth]{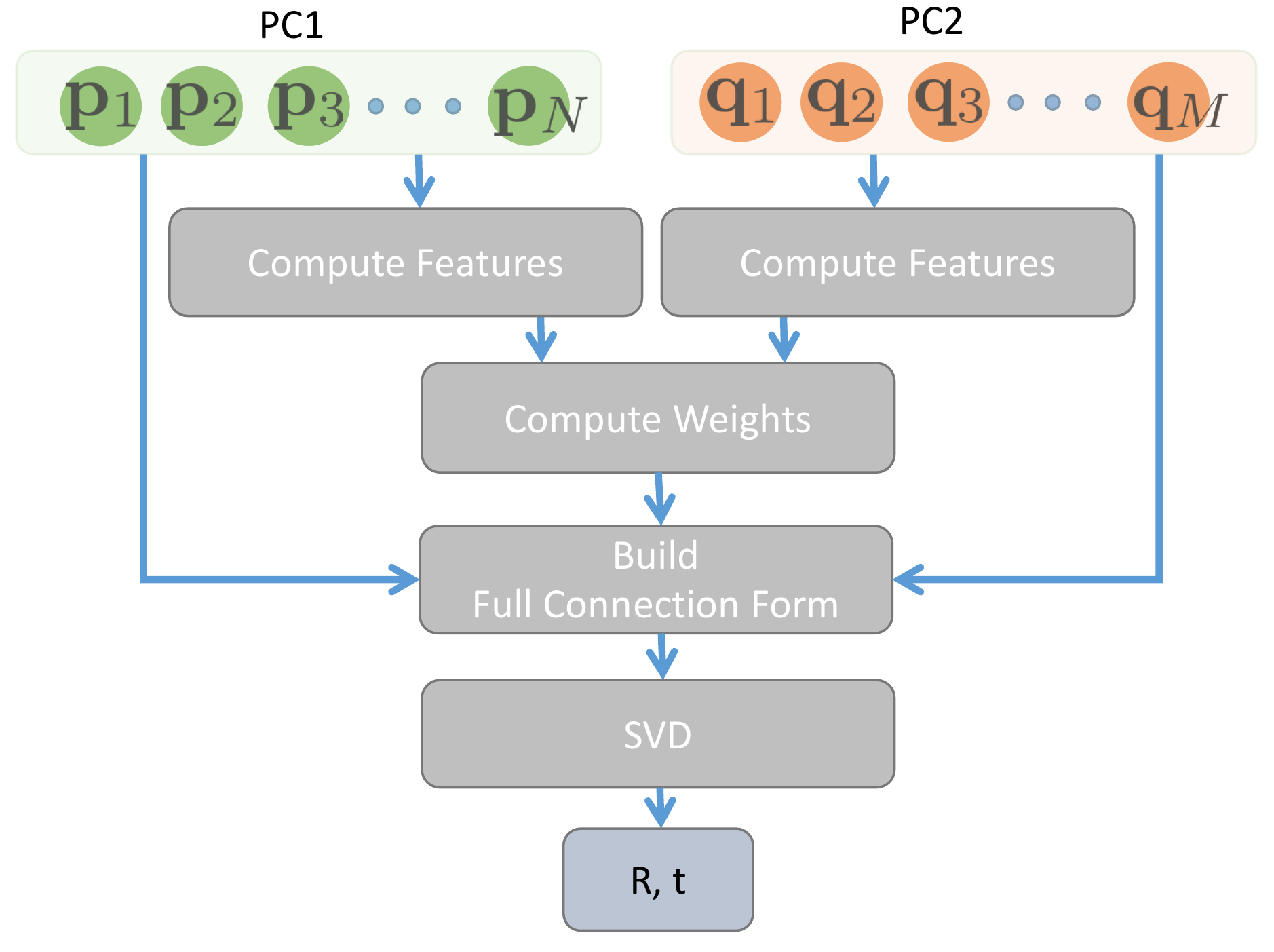}
	\caption{Pipeline of the proposed method.}
	\label{fig:cf_pipe}
\end{figure}
\begin{figure*}[t!]
	\centering
	\subfloat[Small angle, centered]{
		\includegraphics[width=.25\linewidth]{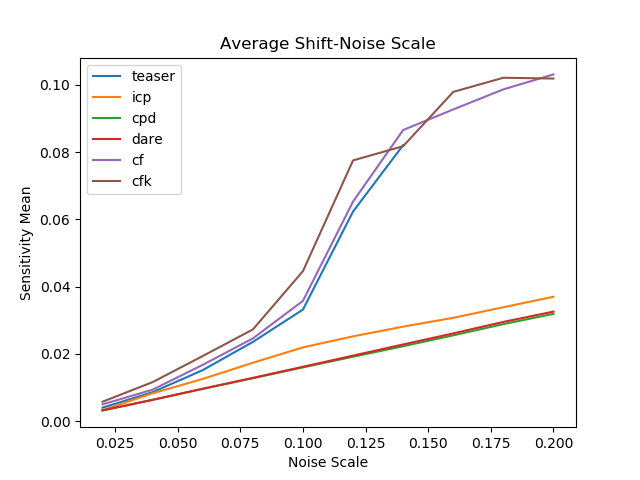}		
		\includegraphics[width=.25\linewidth]{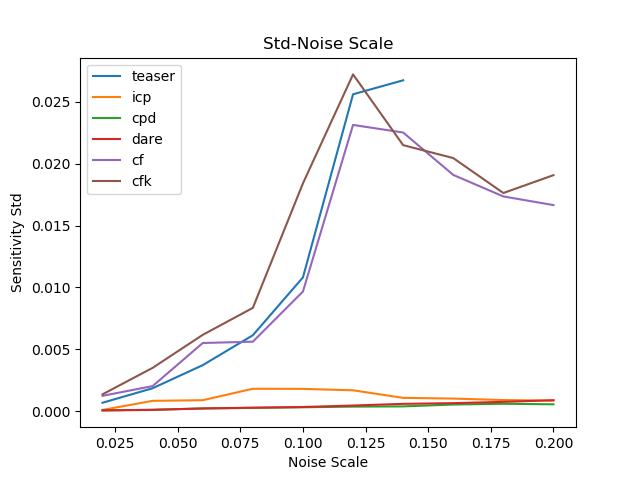}
	}
	\subfloat[Large angle, not centered]{
		\includegraphics[width=.25\linewidth]{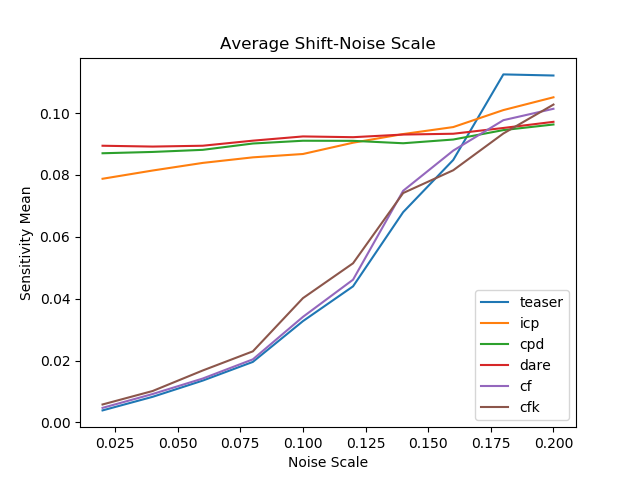}		
		\includegraphics[width=.25\linewidth]{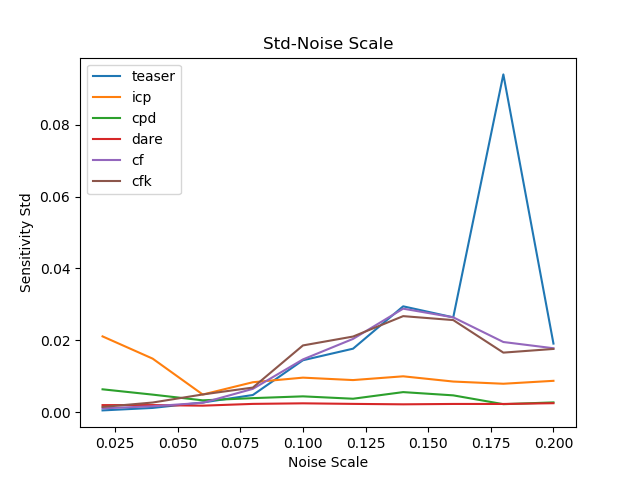}
	}
	\caption{Sensitivity test. The left two plots show results with
		small rotation, centered. The right two plots show results with
		large rotation, not centered. The first and third diagrams show
		the mean shift to noise scale. The second and forth diagrams show
		the standard deviation. }
	\label{fig:averageshift}
\end{figure*}
In addition to $\beta$ and $f$, the feature extraction usually depends
on the chosen radius. This implies performance changes when using
differently scaled data. To make the whole algorithm invariant to
scale, FPFH is using the $k$ nearest neighbor search for normal and feature extraction.
The complete registration pipeline is given in Fig.~\ref{fig:cf_pipe}.

\subsection{Time Complexity}

The runtime for the proposed method is dominated by two parts:
Computing the weights and solving the SVD. For convenience we assume
$M=N$.
To compute the weight, point descriptors of each point cloud are
computed, which takes $\mathcal O(N k \log N)$, where $k$ is the number
of neighbors for each point. Then setting up the $N^2$ weights takes
$\mathcal O(N^2)$.
In the SVD, we first compute the centroid and transform the point
cloud to center, which takes $\mathcal O(N^2)$, because we have to
consider $NM$ terms. Since $W\in \mathbb R^{NM\times NM}$ is a
diagonal matrix, the multiplication for $\M X \M W \M Y^T$ is
equivalent to scaling each row $i$ of $\M Y^T$ with $\M W_{i,i}$.  Thus,
to obtain $\M X \M W \M Y^T$ takes $\mathcal O(N^2)$. As $\M X \M W \M
Y^T$ is a 3-by-3 matrix, solving the SVD costs only constant time.

Overall, the time complexity of proposed method is with $\mathcal
O(N^2)$.

\subsection{A Variant: Applying on Point Set of Keypoints}
For large point sets, the time complexity of $\mathcal O(N^2)$ becomes
infeasible. One possible solution is to extract interest points and
to apply the full connection cost to the two sets of keypoints.

Using FPFH, the implementation is inexpensive. For each point set with
$N$ points, computing the normals takes $\mathcal O(N k \log N)$ and keypoint
detection takes $\mathcal O(N)$. Assume $n$ points are extracted ($n
<< N$), then weight and SVD computation is done on $n$
points. Overall, we yield $max(\mathcal O(N k \log N),\mathcal O(n^2) )$.
\section{Experiments and Results}
\label{sec::exp}

We compare the proposed algorithm with ICP, a feature based state-of-the-art
algorithm (TEASER), Coherent Point Drift (CPD) and
Density Adaptive Point Set Registration (DARE). We call our method
Full Connection Form Solution (CF) and CF-keypoint (CFK) (a variant with keypoints) for short.

\begin{figure}[]
	\vspace{-1cm}
	\centering
	\subfloat[bunny]{
		\includegraphics[width=.3\linewidth]{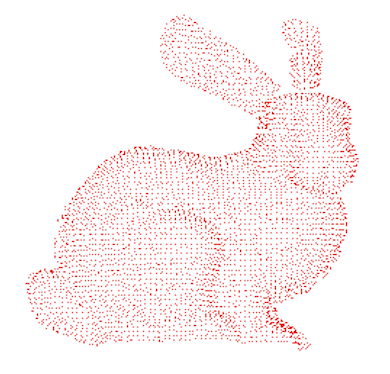}}
	\subfloat[dragon]{
		\includegraphics[width=.3\linewidth]{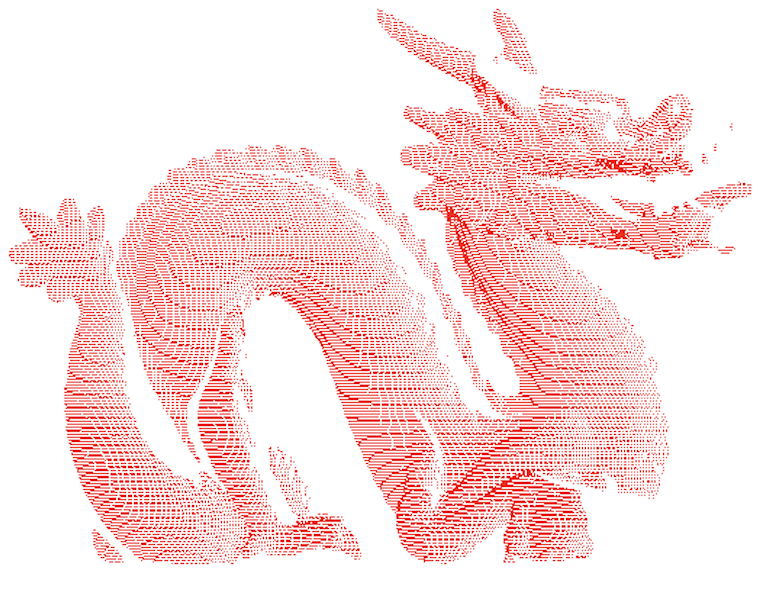}}
	\subfloat[Armadillo]{
		\includegraphics[width=.3\linewidth]{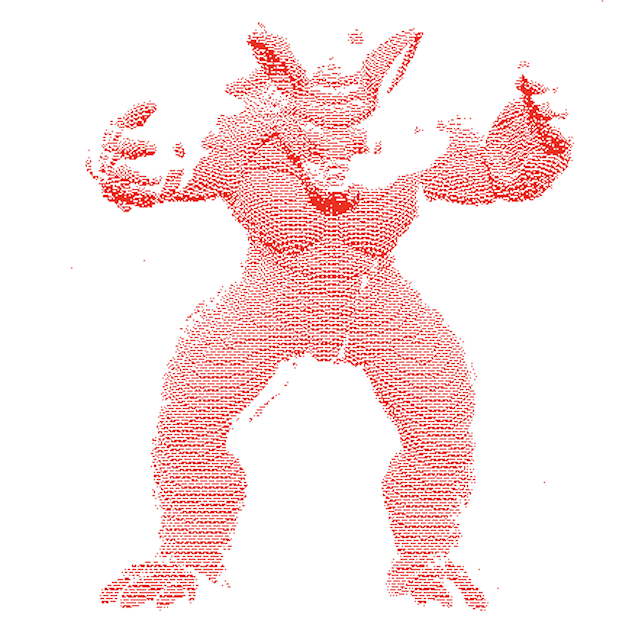}}
	\caption{Three point cloud used for experiments.}
	\label{fig:datum}
\end{figure}

In our experiments, the small 3D object datasets ``bunny'', ``dragon'',
and ``Armadillo'' (bun000, dragonStandRight\_0 and
ArmadilloStand\_180) from the Stanford website\footnote{\url{http://graphics.stanford.edu/data/3Dscanrep/}} have been used. They are in bounding boxes with side lengths ($0.156$, $0.153$, $0.118$), ($0.205$, $0.146$, $0.072$) and ($0.215$, $0.275$, $0.258$) respectively. They are shown in Fig.~\ref{fig:datum}. With those we evaluate our algorithms w.r.t. its sensitivity to noise, the robustness to outliers, and
the accuracy of the registration.
%
%
%
The implementation and test code are given in source files that are
available in github\footnote{Added after acceptance}.
\subsection{Settings}
We first sample the point clouds from the meshes using Meshlab
\cite{cignoni2008meshlab}. 
For CPD the open source C++ implementation from the original project \cite{myronenko2010point} is used. We have set its scale
and reflection parameters to false. For DARE we use the python
implementation of \cite{jaremo2018density}. Its color label and
feature label are disabled.
We also use TEASER from the TEASER++ implementation \cite{yang2020teaser}. 
We have implemented CF and CFK using the Point Cloud Library (PCL)
\cite{rusu2011point}, where we use its FPFH descriptor and the SIFT keypoint detector.
The ICP experiments were also done with PCL. %
The normal and feature computation in CF and CFK are performed
with the same settings, i.e., searching $k$ neighbors. In our
implementation we fixed $k$ to 150. In addition, the $\beta$ used in
Eq.~\eqref{eq:similarity} is fixed to 100.

We set the ICP parameters with max correspondence distance $0.5$, max iteration $1000$, transformation epsilon $1e-9$ and Euclidean fitness epsilon $0.05$.

For TEASER, we use the same settings as for the feature descriptor FPFH. In the matcher of TEASER++, the options absolute\_scale and crosscheck are selected. The solver is using GNC\_TLS with a 1.4 gnc factor, 0.005 rotation cost threshold and 1000 max iterations.

In our experiments, the registration is done using two point clouds
$PC_{a}$ and $PC_{b}$, that were generated with added noise or outliers from the original point cloud, as is described in more detail later.
We then translate and rotate $PC_{b}$ to get $PC_{b}^{'}$. So the
$PC_{a}$ is our PC1 and $PC_{b}^{'}$ is our PC2 and our task is to
align PC1 to PC2 by solving for the transformation. 

In the following experiments $PC_{b}$ is transformed in two distinct ways to generate $PC_{b}^{'}$. Firstly, we apply just a small, random rotation around the point clouds centroid. For the second type of data we apply a large random rotation around the origin of the dataset, which is not the centroid.

The rotation vector is a concise axis-angle representation, for which both the
rotation axis and angle are represented in the same 3-vector. The rotation
angle is the length of this vector.

The small rotation vectors have values drawn uniformly from $[-\pi/8,\pi/8)$, while the large rotation vectors are uniformly drawn from
$[-\pi/2,\pi/2)$. 
%

\subsection{Sensitivity to Noise}
\label{sec::sens}
\begin{figure}[]
	\vspace{-0.8cm}
	\centering
	\subfloat[Small angle, centered]{
		\includegraphics[width=.33\linewidth]{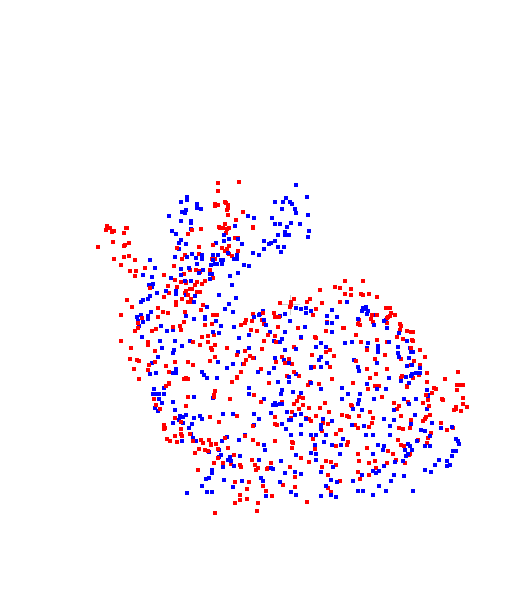}
		\includegraphics[width=.33\linewidth]{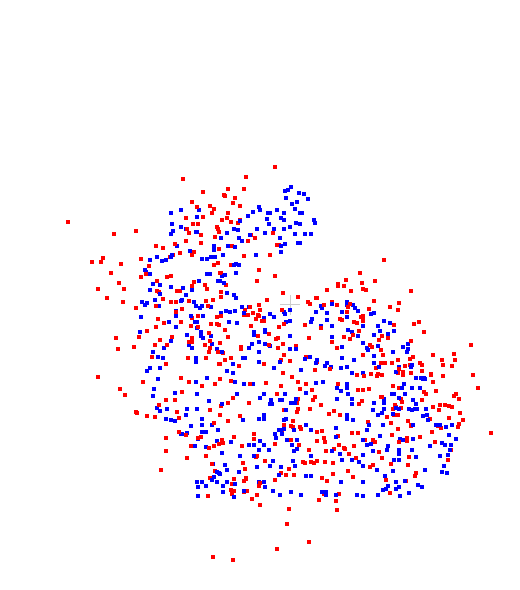}
		\includegraphics[width=.33\linewidth]{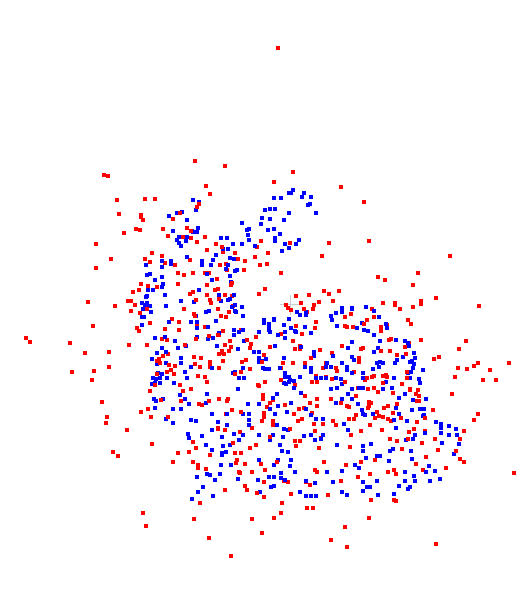}
	}\\
	\subfloat[Large angle, not centered]{
		\includegraphics[width=.33\linewidth]{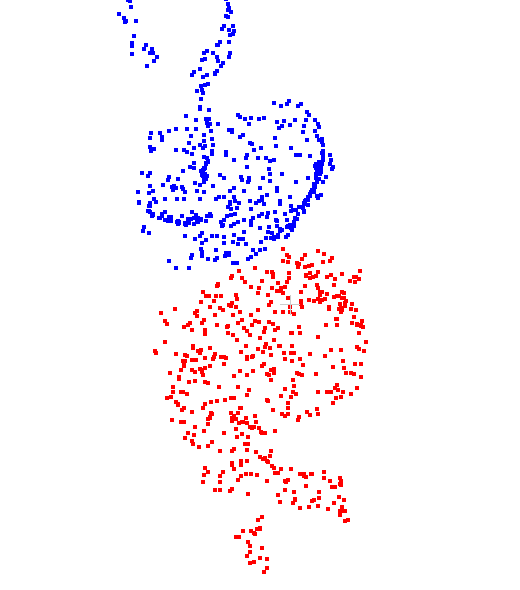}
		\includegraphics[width=.33\linewidth]{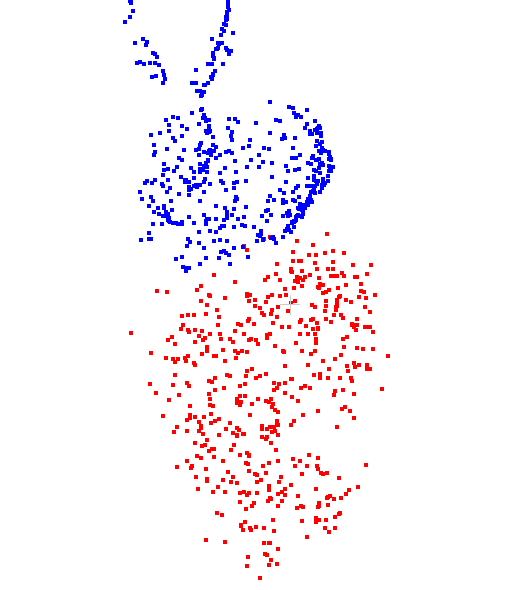}
		\includegraphics[width=.33\linewidth]{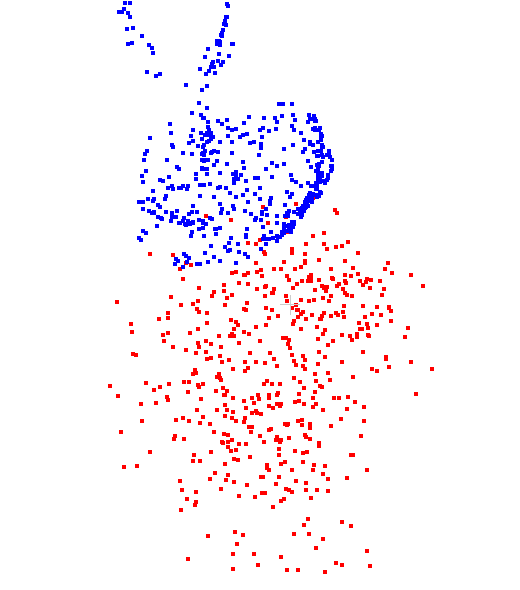}
	}
	\caption{Noise data. Above: centered small angle, below: large
		angle. From left to right column is with noise standard
		derivation 0.002, 0.01 and 0.02.}
	\label{fig:noisedata}
\end{figure}
\begin{table}[b!]
	\centering
	\caption{Robustness test: smaller is better.}
	\label{tab::robust}
	\scalebox{0.85}{
		\begin{tabular}{|l| l | l |}
			\hline
			& Small rotation, centered  & Large rotation, not centered \\ \hline \hline
			ICP&$0.0019\pm 0.0062$ & $0.070 \pm 0.023$\\ \hline
			CPD&$2.4e-09\pm 1.7e-10$&$0.040\pm 0.047$ \\ \hline
			DARE& $0.012 \pm 0.019$&$0.053\pm 0.035$ \\ \hline
			TEASER&$0.0055\pm0.0027$&$0.0057\pm0.0035$ \\ \hline
			\textbf{CF}&$0.0075\pm 0.0028$& $0.0078\pm 0.0032$ \\ \hline
			CFK&$0.0096\pm0.0043$& $0.0099\pm 0.0052$\\\hline
		\end{tabular}}
\end{table}
\begin{figure*}[t!]
	\centering
	\subfloat[Initial]{
		\includegraphics[width=.13\linewidth]{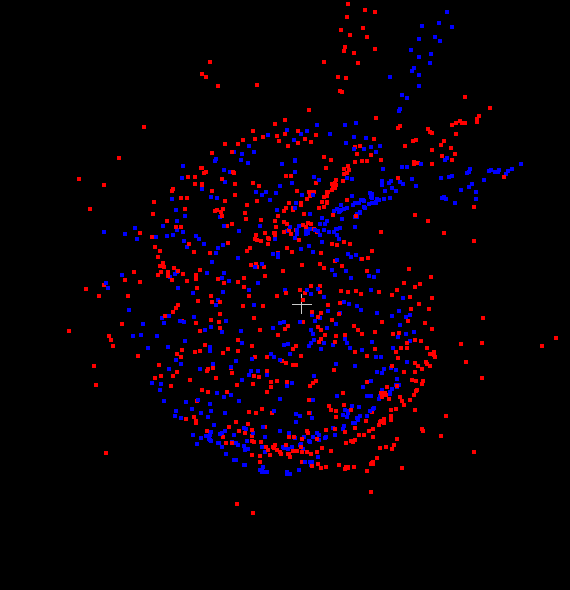}
	}
	\subfloat[ICP]{
		\includegraphics[width=.13\linewidth]{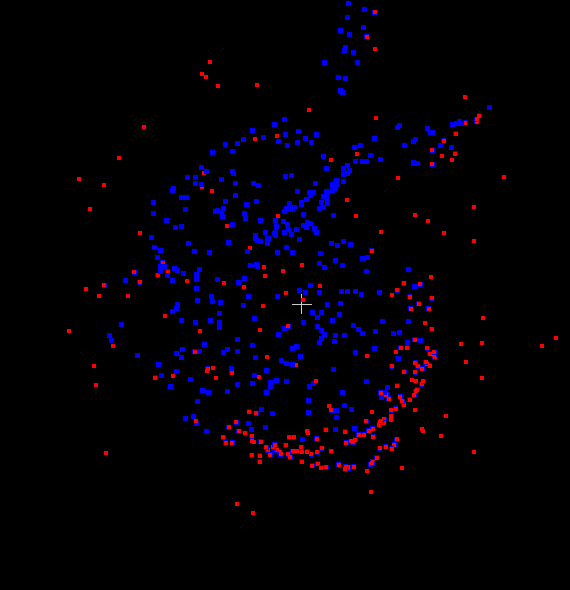}
	}	
	\subfloat[CPD]{
		\includegraphics[width=.13\linewidth]{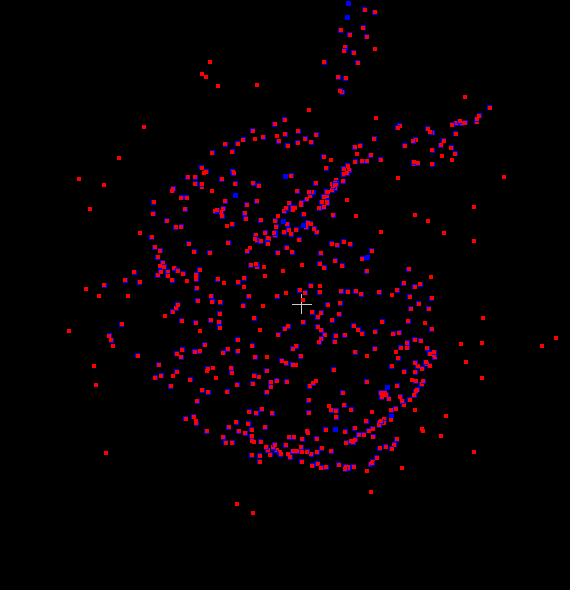}
	}	
	\subfloat[DARE]{
		\includegraphics[width=.13\linewidth]{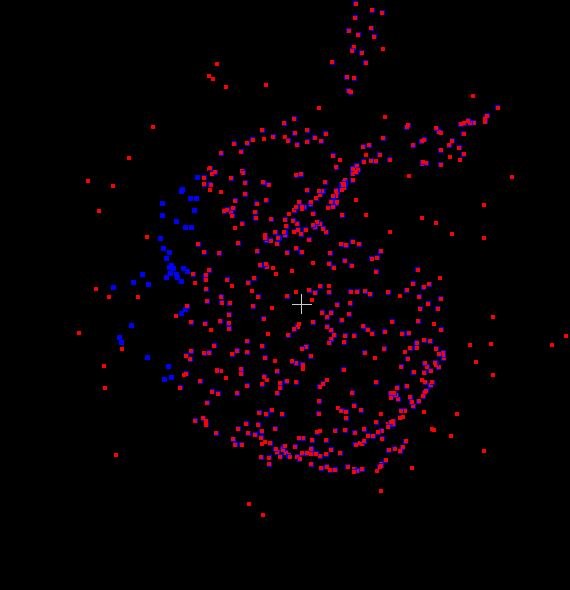}
	}	
	\subfloat[TEASER]{
		\includegraphics[width=.13\linewidth]{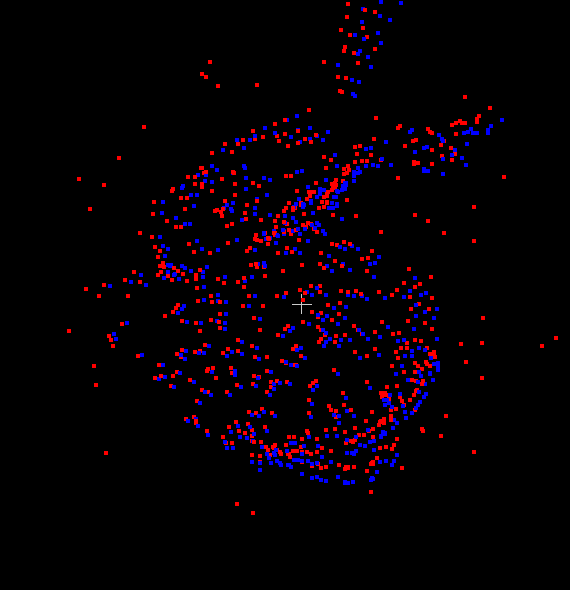}
	}	
	\subfloat[CF]{
		\includegraphics[width=.13\linewidth]{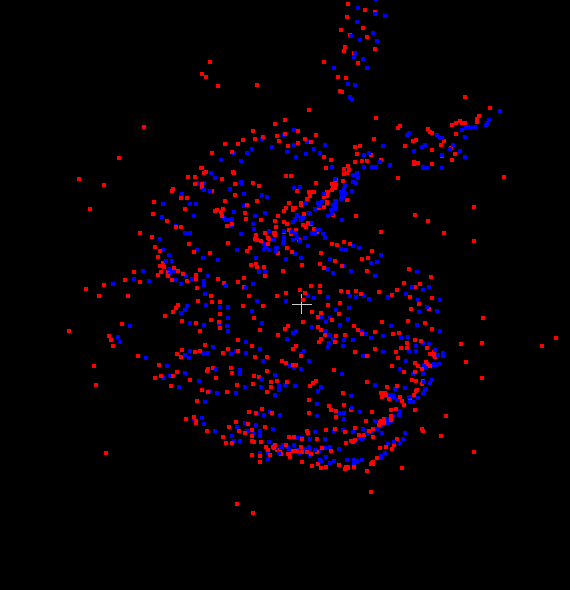}
	}
	\subfloat[CFK]{
		\includegraphics[width=.13\linewidth]{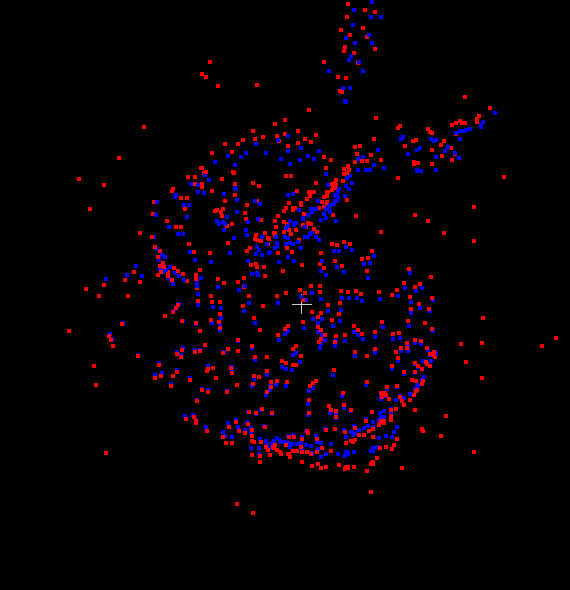}
	}		
	\caption{Robustness test: Example from the small rotation set, centered.}
	\label{fig:outlier_small}
\end{figure*}
\begin{figure*}[t!]
	\vspace{-0.7cm}
	\centering
	\subfloat[Initial]{
		\includegraphics[width=.13\linewidth]{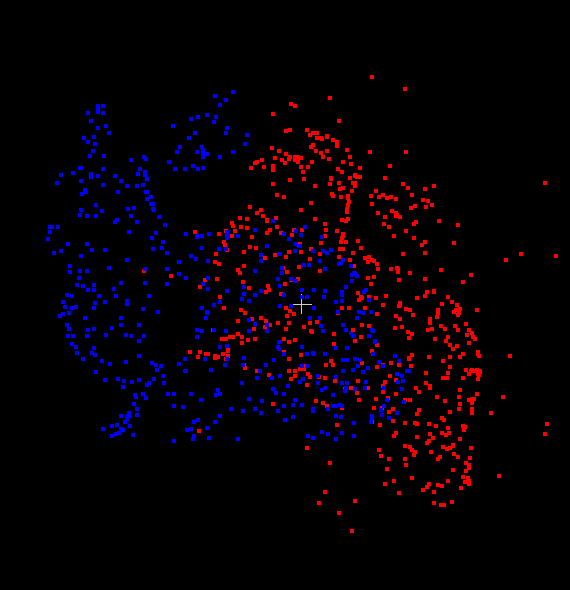}
	}
	\subfloat[ICP]{
		\includegraphics[width=.13\linewidth]{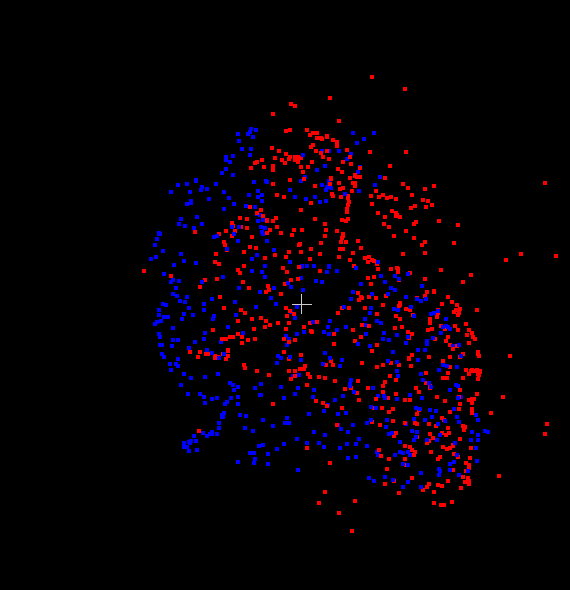}
	}	
	\subfloat[CPD]{
		\includegraphics[width=.13\linewidth]{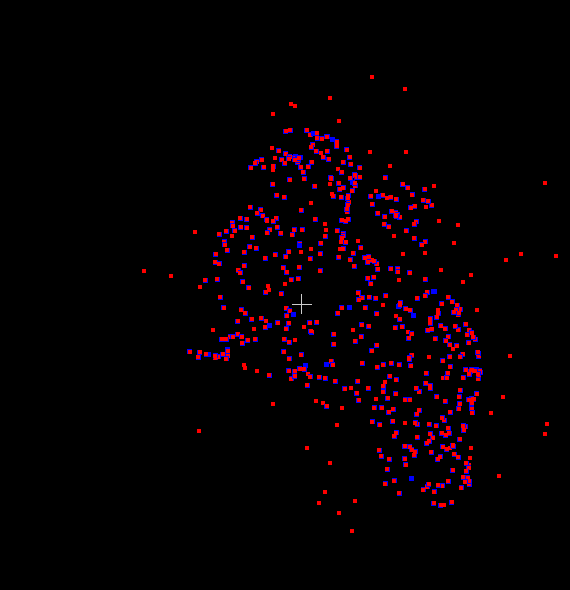}
	}	
	\subfloat[DARE]{
		\includegraphics[width=.13\linewidth]{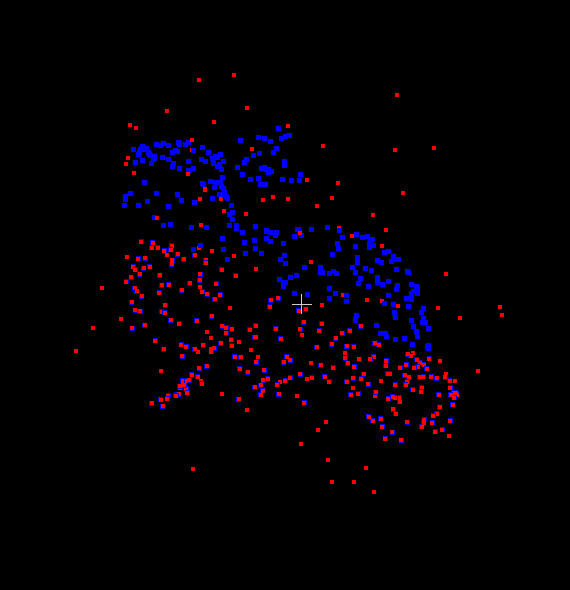}
	}	
	\subfloat[TEASER]{
		\includegraphics[width=.13\linewidth]{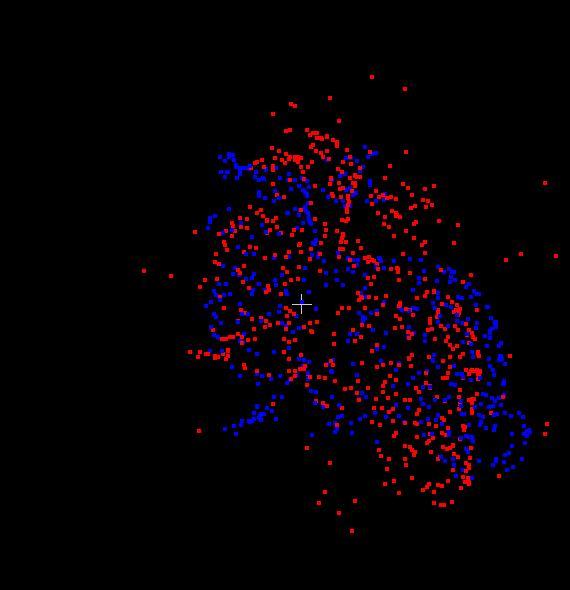}
	}	
	\subfloat[CF]{
		\includegraphics[width=.13\linewidth]{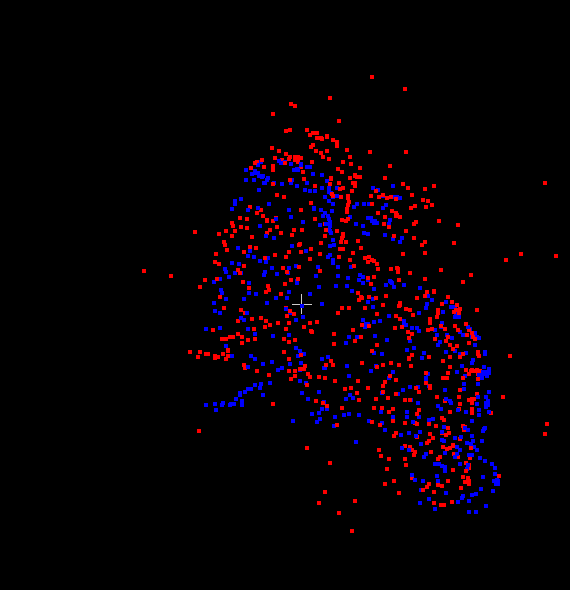}
	}
	\subfloat[CFK]{
		\includegraphics[width=.13\linewidth]{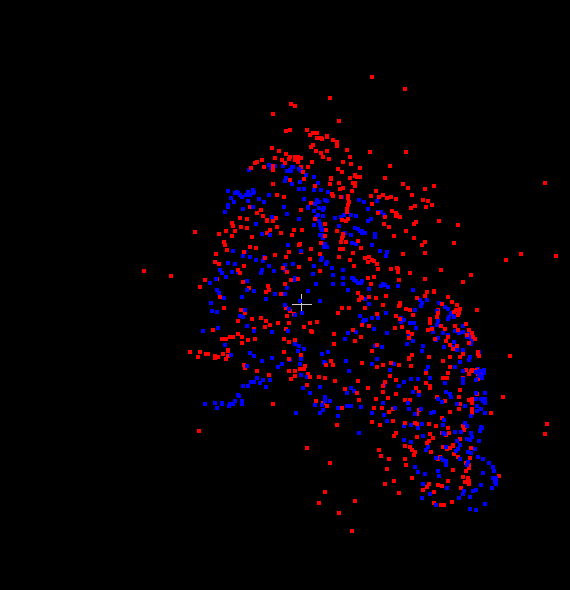}
	}		
	\caption{Robustness test: Medium rotation example from the large rotation set, not centered.}
	\label{fig:outlier_large}
\end{figure*}
\begin{figure*}[t!]
	\vspace{-0.7cm}
	\centering
	\subfloat[Initial]{
		\includegraphics[width=.13\linewidth]{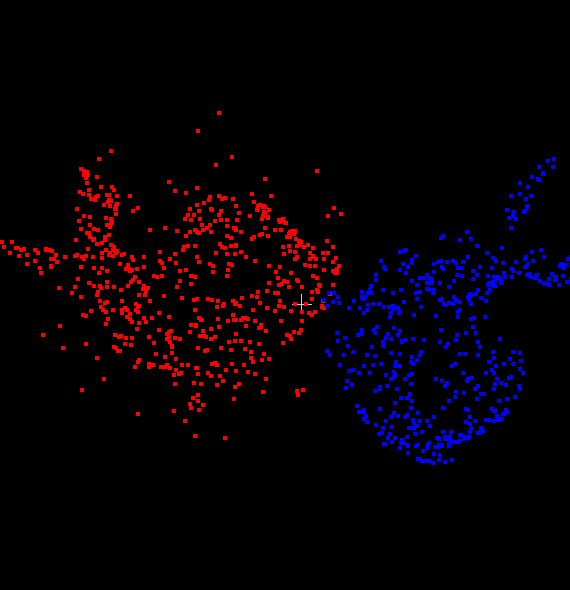}
	}
	\subfloat[ICP]{
		\includegraphics[width=.13\linewidth]{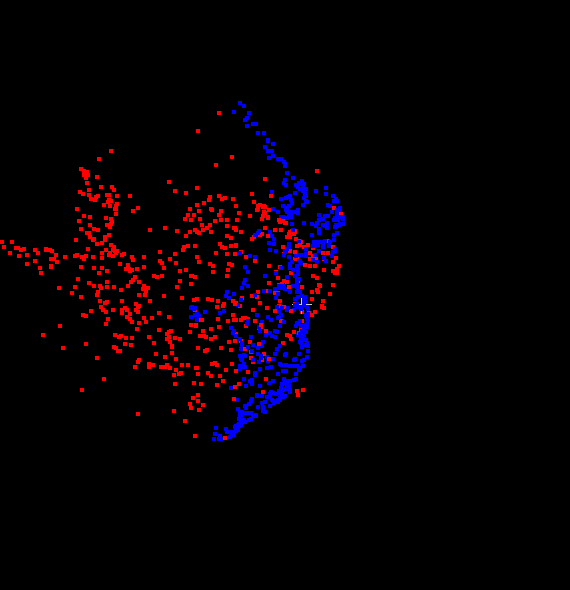}
	}	
	\subfloat[CPD]{
		\includegraphics[width=.13\linewidth]{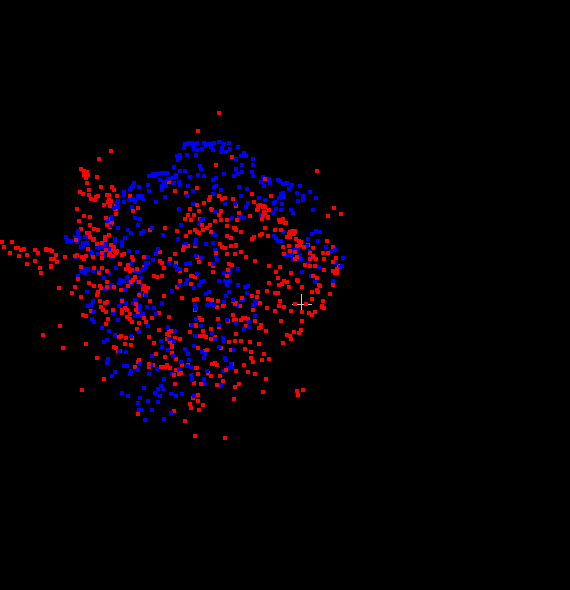}
	}	
	\subfloat[DARE]{
		\includegraphics[width=.13\linewidth]{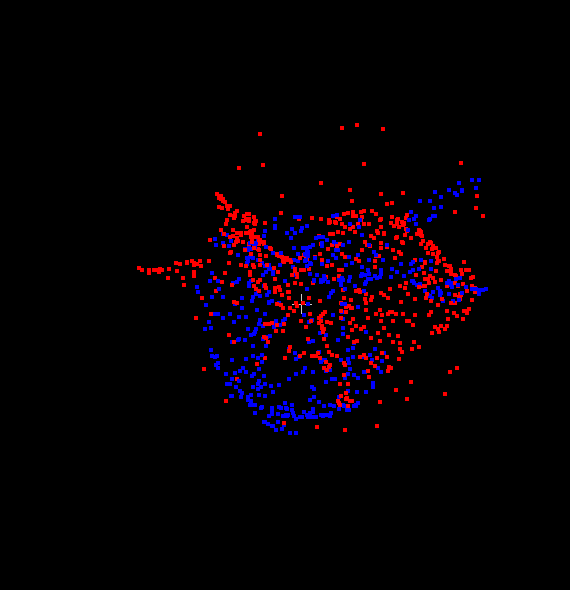}
	}	
	\subfloat[TEASER]{
		\includegraphics[width=.13\linewidth]{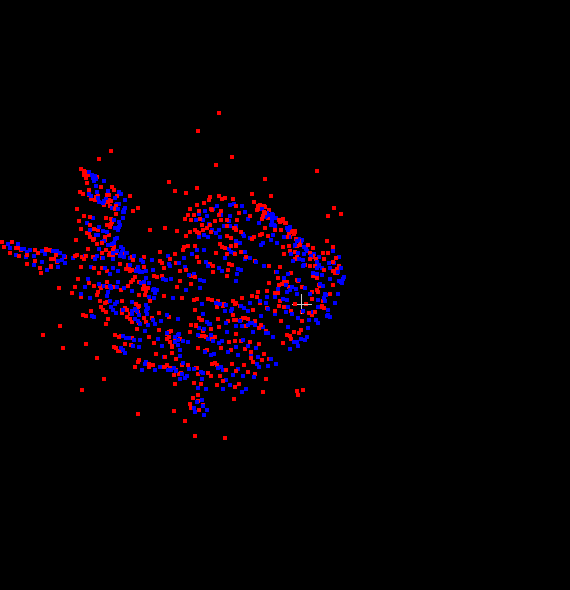}
	}	
	\subfloat[CF]{
		\includegraphics[width=.13\linewidth]{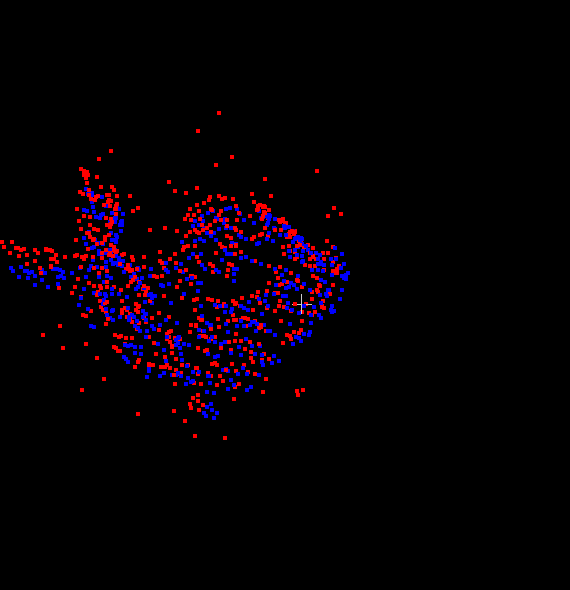}
	}
	\subfloat[CFK]{
		\includegraphics[width=.13\linewidth]{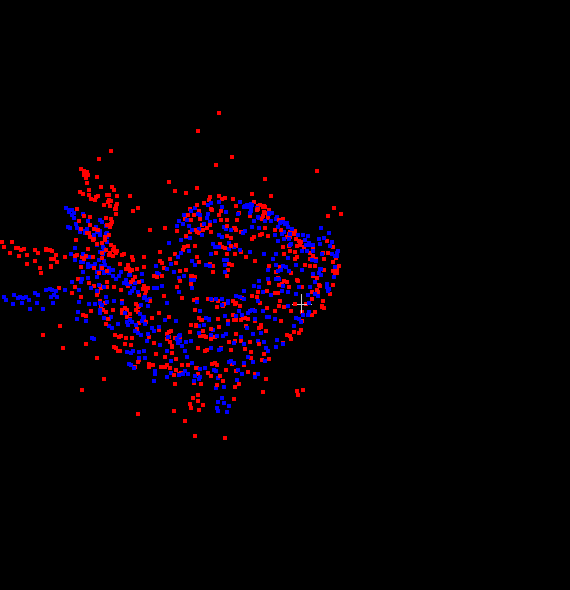}
	}		
	\caption{Robustness test: Large rotation example from the large rotation set, not centered.}
	\label{fig:outlier_large4}
\end{figure*}
In this experiment we evaluate the effects of different levels of noise on the registration. Each level is tested with 30 generated point clouds. Just for this experiment, we fix the large and small rotation angle to two certain values, to be able to concentrate on the effects of the levels of noise and draw the  diagrams of Fig.~\ref{fig:averageshift}.
PC1 and PC2 are subsampled to 500 points. For the rotated set PC2, we
add zero mean Gaussian noise to each point. 


Following the definition of
sensitivity~\cite{tsin2004correlation}, we log the mean average shift
to evaluate the performance and the standard deviation is utilized as the
metric.
The noise scale is within the range $(0,0.02]$. Because the size of the
bunny does not exceed 0.3, too large noise will result in
dysfunctional feature descriptors.
We present the noise data with different noise scale in
Fig.~\ref{fig:noisedata}.
The results are given in Fig.~\ref{fig:averageshift}. In the small
angle case of Fig.~\ref{fig:averageshift}, the TEASER curve breaks
due to a low number of correspondences and followed by failure. 

For the centered small rotation, ICP, CPD and DARE achieve better
average shifts and less sensitivity to noise. For the feature
based methods, our CF and CFK perform very similar to TEASER. 

However, for the large rotation data, ICP, CPD and DARE fail to
align the point clouds, while the feature based methods CF, CFK, and
TEASER are able to align with good performance. 

\begin{table*}[t]
	\centering
	\caption{Accuracy test: smaller is better.}
	\label{tab::acc}
	\scalebox{0.85}{
		\begin{tabular}{ | l| l | l | l |     l| l| l| }
			\hline
			&\multicolumn{3}{l|}{Small rotation, centered} &\multicolumn{3}{l|}{Large rotation, not centered} \\ \hline
			&bunny &dragon&Armadillo &bunny &dragon&Armadillo\\ \hline  \hline
			
			ICP& $0.045\pm0.018$ & $0.045\pm 0.19$ & $ 0.036\pm 0.022$& 
			$0.96\pm 1.14$ & $1.09\pm 1.13$ &  $1.17\pm 1.15$\\ \hline
			CPD&$0.016\pm 0.0089$  &$0.014\pm 0.0083$  &$0.012\pm 0.0074$ &
			$1.15\pm 1.30$&$1.11\pm 1.19$& $1.13\pm 1.15$\\ \hline
			DARE &$0.020\pm 0.0093$ & $0.016\pm 0.0090$& $0.016\pm 0.0083$&
			$1.30\pm 1.26$& $1.34 \pm 1.22$ & $1.48 \pm 1.16$ \\ \hline
			TEASER &$0.14\pm 0.076$ &$0.15\pm 0.084$&$0.16\pm 0.095$ &
			$0.15\pm 0.096$&$0.13\pm 0.082$&$0.14\pm 0.093$\\ \hline
			\textbf{CF}& $0.16\pm 0.10$ & $0.19\pm 0.13$ & $0.28\pm 0.42$&
			$0.18\pm0.14$ & $0.14\pm 0.11$ & $0.15\pm 0.12$\\ \hline
			CFK& $0.26\pm 0.26$ &  $0.25\pm 0.22$ & $0.33\pm 0.35$&
			$0.26\pm 0.23$ & $0.19\pm 0.18$ & $0.20\pm 0.17$\\ \hline
			
		\end{tabular}}
	\end{table*}

\subsection{Robustness to Outliers}
\label{sec::outlier}
Similarly, we also use 500 randomly selected points from the bunny object
and perform small and large rotations. Additionally, 100 random points have
been uniformly drawn in a spherical way and added to the rotated point
set PC2 (with radius 0.2, around the center of sampled point clouds).

Because the first 500 points in each set are also from the same
sampled index, we actually know the correspondence in the non-outlier
parts. To quantify the robustness, we compute the average shift as in
subsection~\ref{sec::sens}.

For both large and small rotations, we test 100 times to record the
mean and standard deviation. The quantitive evaluation is given in
Table~\ref{tab::robust}. Selected visualizations of the alignment are
presented in Fig.~\ref{fig:outlier_small},
Fig.~\ref{fig:outlier_large}, and Fig.~\ref{fig:outlier_large4}. All
experiments have been made using randomly drawn rotation vectors.

CPD achieves extremely precise solutions for small rotations, while
feature based methods (TEASER, CF, CFK) are similar and are better than ICP
and DARE. DARE gives the largest error and standard deviation. For
the large rotation case, the feature based methods (TEASER, CF, CFK)
perform best and the errors of the remaining methods are several times
worse and unstable, since they yield large standard deviations. The performance of our one-step methods are close to TEASER, even though its truncated least square is theoretically more insensitive to spurious data.
%

We select tests with one centered small rotation and two large
rotations for demonstration. In Fig.~\ref{fig:outlier_small} and
Fig.~\ref{fig:outlier_large}, CPD and DARE achieve better alignment,
while they fail in Fig.~\ref{fig:outlier_large4}. However, our CF and
CFK keep similar results, but are affected by the outliers.


\subsection{Accuracy}

Using the same given transformation applied to the original point sets
as in subsection~\ref{sec::outlier}, we achieve rotated models. Then
we randomly sample 500 points from both the reference model and the
rotated models for testing. In the accuracy test, the three point sets in Fig.~\ref{fig:datum} (bunny, dragon and Armadillo) are utilized.
To evaluate the accuracy, deviations from the identity matrix
\cite{larochelle2007distance} are computed:
\[ACC_{\M R_\text{gt}}(\M R_\text{predicted}) =  ||\M I - \M R_\text{predicted}\M R_\text{gt}^T||_F \]
It is a distance measure using the Frobenius norm of a matrix, where
$\M R_\text{gt}$ is the given rotation and $\M R_\text{predicted}$ is the
predicted rotation.

Accuracy results are given in Table~\ref{tab::acc}. For the centered
small rotation case, we observe that CPD also achieves the best score
while TEASER, CF and CFK are on the same level. For large rotations,
CPD becomes unstable, which results in much larger average rotation
distances and their standard deviations. The feature based methods
still show close results in different cases. Our one-step solution achieves similar result to the truncated least square method TEASER.

%
%





\section{Discussion}

From the experiments, we find that the feature based algorithms (TEASER, CF,
CFK) are more sensitive to noise that the other approaches. Though with very small initial
rotation, the noise still has impact on the result. In the outliers
test, CPD yields very precise results. The one-step CF algorithm scores close to state-of-the-art TEASER.

However, we also test with large rotation angles and observe, that the
feature based algorithms are not affected, which is shown by their
very close results in the curves and tables, while the remaining
methods fail or have large errors.

\begin{figure}[b!]
	\centering
	\subfloat[Initial state.]{
		\includegraphics[width=.49\linewidth]{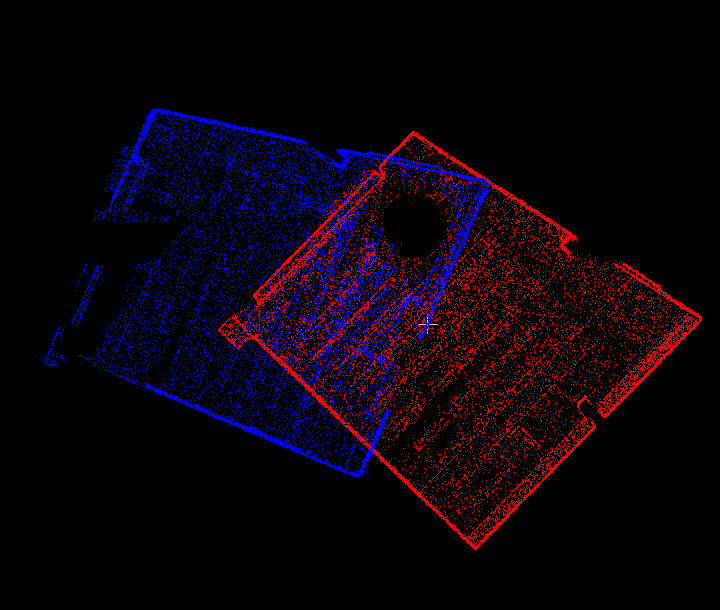}
		\label{fig:lecturehall}
	}
	\subfloat[Align two frames.]{
		\includegraphics[width=.49\linewidth]{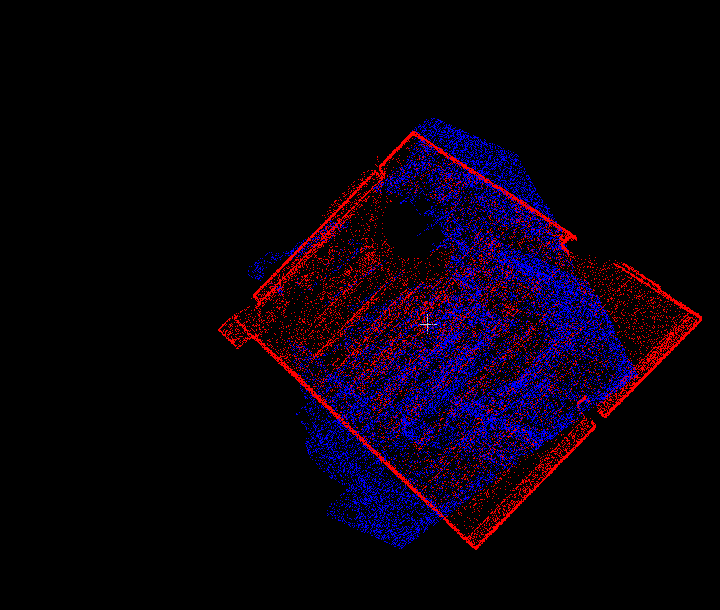}
		\label{fig:lecturehall_aligned}
	}
	\caption{Align the Lecture hall data with partial overlap.}
	\label{fig:wuz}
\end{figure}

We also try our method on the Lecture Hall dataset\footnote{\url{http://kos.informatik.uni-osnabrueck.de/3Dscans/}}. 
It is a small lecture hall with approx 60 seats and it was scanned from two vantage points close to corners by a high-end Riegl VZ-400 laser scanner with an angular resolution horizontal and vertical of 0.04 deg.
One of the scans has two people in it, holding a blanket, while the other does not, thus the scene wasn't static.
After sub-sampling, we select 30,000 points randomly from each of the two scans, which are shown in Fig.\ref{fig:lecturehall}. We apply CFK on the data
and it fails to aligns the point cloud as seen in Fig.~\ref{fig:lecturehall_aligned}. In our previous experiment, the algorithm is capable to align point clouds from the same distribution. While in such lecture hall case, the missing data on its corresponding region of the other induce additional error to the one-step solved solution. However, trimmed correspondence and truncated loss are able to solve such case iteratively. Thus, our following work will focus on the distinctiveness of descriptors and more complicated weight designation on this one-step solution.
This way the method will be applicable to various kinds of data.
\section{Conclusions}
\label{sec::conclude}
In this paper we have proposed a new solution to the point set registration
problem that does not require correspondences. From our survey, this is the
first algorithm that directly provides accurate registrations in a non-iterative
one-step setup, where the derived formulas have closed-form solution. Future
work will focus on investigating different feature descriptors and weights to
achieve better robustness for non-identical distribution data. 
The experiments in
this paper demonstrate competing performance to various methods in robustness to
outliers and accuracy. The feature based algorithms (TEASER, CF, CFK) are more
sensitive to noise compared to the non-feature based methods due to
defunctioning descriptors. Overall, our one-step solution, CF and CFK provide close performance to state-of-the-art TEASER with respect to noise, outliers and accuracy. The non-feature based methods fail or give much worse
results while our methods are more stable at large rotations and thus achieve
better scores and standard deviations. 

\begin{small}
  \section*{Acknowledgements}
  The authors would like to thank Heng Yang and Luca Carlone from
  Massachusetts Institute of Technology for providing the TEASER code
  and discussing the topic. This work was supported by a German
  Academic Exchange Service (DAAD) scholarship granted to Yijun Yuan.
\end{small}  

\pagebreak
\IEEEtriggeratref{12}
\bibliographystyle{IEEEtran}
\bibliography{reference}

\end{document}